\documentclass[
]{ceurart}
\pdfoutput=1
\usepackage{lineno,hyperref}

\usepackage[utf8]{inputenc}
\usepackage{graphics}
\usepackage{dsfont}
\usepackage{caption}
\usepackage{amsmath,mathtools}
\usepackage{newunicodechar}
\usepackage{xspace}
\usepackage{enumitem}
\usepackage{hyperref}
\usepackage{multicol}
\usepackage{multirow}
\usepackage{tabu}
\usepackage{booktabs}
\usepackage{array}%
\usepackage{esvect}
\usepackage{makecell}

\usepackage{url}
\usepackage{tabularx}
\usepackage{xurl}
\usepackage{adjustbox}

\usepackage{xcolor}
\usepackage{appendix}
\usepackage{longtable}

\usepackage{marginnote}

\usepackage{listings}
\begin{document}

\copyrightyear{2022}
\copyrightclause{Copyright for this paper by its authors.
  Use permitted under Creative Commons License Attribution 4.0
  International (CC BY 4.0).}

\conference{Please cite Colla D., Delsanto M., Agosto M., Vitiello B., Radicioni D. P. (2022). Semantic coherence markers: The contribution of perplexity metrics. Artificial Intelligence in Medicine, 134, 102393.\newline \url{https://doi.org/10.1016/j.artmed.2022.102393},~\cite{colla2022semantic}}
\title{Semantic Coherence Markers\\ for the Early Diagnosis of the Alzheimer Disease}




\author[1]{Davide Colla}[%
email=davide.colla@unito.it
]
\address[1]{Dipartimento di Informatica, Università degli Studi di Torino}

\author[1]{Matteo Delsanto}[%
email=matteo.delsanto@unito.it
]

\author[2]{Marco Agosto}[%
email=marco.agosto@unito.it
]
\address[2]{Dipartimento di Scienze della Sanità Pubblica e Pediatriche, Università degli Studi di Torino}

\author[2]{Benedetto Vitiello}[%
email=benedetto.vitiello@unito.it
]

\author[1]{Daniele P. Radicioni}[%
email=daniele.radicioni@unito.it
]


\begin{abstract}
%
In this work we explore how language models 
can be employed to analyze language and discriminate between mentally impaired and healthy subjects through the perplexity metric.
%
Perplexity was originally conceived as an information-theoretic measure to assess how much a given language model is suited to predict a text sequence or, equivalently, how much a word sequence fits into a specific language model. 
We carried out an extensive experimentation with the publicly available data, and employed language models as diverse as N-grams ---from 2-grams to 5-grams--- and GPT-$2$, a transformer-based language model. 
%
%
%
We investigated whether perplexity scores may be used 
to discriminate between the transcripts of healthy subjects and subjects suffering from Alzheimer Disease (AD). Our best performing models achieved full accuracy and F-score ($1.00$ in both precision/specificity and recall/sensitivity) 
in categorizing subjects from both the AD class and 
control subjects.
These results suggest that perplexity can be a valuable analytical metrics with potential application to supporting early diagnosis of symptoms of mental disorders. 

\end{abstract}


\maketitle
\section{Introduction}
%
%

{\footnotesize
\begin{quote}
\textit{This paper is the (significantly) abridged version of the article ``Semantic coherence markers: The contribution of perplexity metrics'' (\url{https://doi.org/10.1016/j.artmed.2022.102393},~\cite{colla2022semantic}), which also contains references to employed data and to the implementation of the described work.}
\end{quote}
}

\noindent In economically developed societies the burden of mental disturbances is becoming more evident, with negative impact on people's daily life and huge cost for health systems. Whereas for many psychotic disorders no cures have been found yet, the treatment of people at high risk for developing schizophrenia or related psychotic disorders is acknowledged to benefit from early detection and intervention~\cite{marshall2005association}. 
To this end, a central role might be played by approaches aimed at analyzing thought and communication patterns in order to identify early symptoms of mental disorder~\cite{larson2010early}.

The analysis of human language has recently emerged as a research field that may be helpful to analyze for diagnosing and treating mental illnesses. 
%
Recent advances in NLP technologies allow accurate language models (LMs) to be developed. These can be thought of as probability distributions over text sequences, that can be used to estimate in how far a text is coherent with (or, more precisely, predictable through) such language models. In order to measure the distance between an actual sequence of tokens and the probability distribution we propose using \emph{perplexity}, a metric that is well-known in literature for the intrinsic evaluation of LMs. 
In this work we 
report results on a simple experiment, aimed at assessing 
whether the perplexity 
can help in discriminating healthy subjects from people suffering from mental disorders.

Although in literature perplexity is not new as a tool to compare the language of healthy and diagnosed subjects, we report experimental results favorably comparing with those in literature. 
Moreover, as far as we know, no previous work has compared perplexity scores computed through LMs as diverse as GPT-$2$ and N-grams to the ends of discriminating healthy subjects from subjects afflicted by Alzheimer Disease. This difference has practical consequences for applications, mostly due to the different computational effort required both to train and employ such models, and to the descriptive power of the learned models. 
\section{Related Work}\label{sec:related_work}
%

%
In the last decade, advances in NLP techniques have allowed the construction of approaches to automatically deal with tasks such as linguistic analysis and production, including also many of the aforementioned linguistic levels.
These approaches have identified markers that can help differentiate patients with psychiatric disorders from healthy controls, and predict the onset of psychiatric disturbances in high risk groups at the level of the individual patient.

Although originally conceived to assess how language models are able to model previously unseen data, perplexity can be used to compare (and discriminate) text sequences produced by healthy subjects or by people suffering from language-related disturbances. To provide a hint of this approach, perplexity is a positive number that ---given a language model and a word sequence--- expresses how unlikely it is for the model to generate that given sequence. A richer description of the perplexity is provided in Section~\ref{sec:background}.
In~\cite{stolcke1996statistical} N-grams of part of speech (POS) tags were employed to identify patterns at the syntactic level. Then, two LMs were acquired (one from patients' data and the other from data from healthy controls): the categorization of a new, unseen (that is, not belonging to either set of training data) sample was then performed through the perplexity computed with the two LMs over the sample. 
%
The considered sample was then categorized as produced by a healthy subject (patient) if the LM acquired from healthy subjects (patients) data attained smaller perplexity than the other language model.
%
%
%
Perplexity has been recently proposed as an indicator of cognitive deterioration~\cite{frankenberg2019perplexity}; more specifically, the content complexity in spoken language has been recorded in physiological aging and at the onset of Alzheimer’s disease (AD) and mild cognitive impairment (MCI) on the basis of interview transcripts. 
LMs used in this research were built by exploiting 1-grams and 2-grams information; as illustrated in next section (please refer to Equation~\ref{eq:w_seq_prob_approx}), such models differ in the amount of surrounding information employed. 
Perplexity scores were computed on ten-fold-cross-validation basis, whereby  participants' transcripts were partitioned into ten parts; a model was then built by using nine parts and was tested on the tenth. This procedure was repeated ten times so that each portion of text was used exactly once as the test set.
Four examination waves with an observation interval of more than 20 years were performed, and correlations of the perplexity score of transcriptions dating to the beginning of the experiment were found with the score from the dementia screening instrument in participants that lately developed MCI/AD.

Perplexity has been employed as a predictor for Alzheimer Disease (AD) on the analysis of transcriptions from DementiaBank's Pitt Corpus, that contains data from both healthy controls and AD patients~\cite{becker1994natural}. More precisely, in~\cite{fritsch2019automatic} two neural language models, based on LSTM models, were acquired, one built on the healthy controls and the other trained on patients belonging to the dementia group. A leave-one-speaker-out cross-validation was devised and, according to this setting, a language model $\mathcal{M}_{-s}$ was created for each speaker \textit{s} by using all transcripts from the speaker's group but those of \textit{s}. Data from speaker \textit{s} was then tested on both $\mathcal{M}_{-s}$, thus providing a perplexity score $p_{own}$, and on the language model built upon the transcripts from the whole group to which the speaker did not belong to, thus obtaining the perplexity score $p_{other}$. The difference between the perplexity scores $\Delta_{s} = p_{own} - p_{other}$ was computed
as a description for the speaker \textit{s}.
The classification of each speaker was then performed by setting a threshold ensuring that both groups obtained equal error rate.
The authors achieved $85.6\%$ accuracy on $499$ transcriptions, and showed that perplexity can also be exploited to predict a patient's Mini-Mental State Examination (MMSE) scores. 
%
%
%
The approach adopted in this work is the closest to our own work we could find in literature; however it also differs from ours in some aspects. First, we investigated how reliable perplexity is in assessing the language of healthy subjects. That is, we analyzed how perplexity scores vary within the same individual, as an initial step toward assessing if perplexity is suitable for examining text excerpts/transcripts that (like in the case of the Pitt Corpus) were collected through multiple interviews and tests, spanning over years.
Additionally, we were concerned with evaluating all excerpts from a single individual to predict the AD diagnosis at the subject level, rather than in predicting the class for each and every transcript. In order to assess the perplexity as a tool to support the diagnosis, we analyzed only data from subjects for which at least two transcripts were available.

Following the approach presented in~\cite{fritsch2019automatic},  perplexity has been further investigated for the categorization of healthy subjects and AD patients~\cite{cohen2020tale}. In particular, different LMs have been acquired on both control and AD subjects' transcriptions from the Pitt Corpus~\cite{becker1994natural}. Such LMs have been employed to evaluate in how far differences in perplexity scores reflect deficits in language use.
%
%
%
Our approach differs from this one. Firstly, we explored two different sorts of LMs (N-grams and GPT-2 models, fine tuned with $5$, $10$, $20$ and $30$ epochs) so to collect experimental evidence on the level of accuracy recorded by different LMs used to compute the perplexity scores. Secondly, four different decision rules were compared based on average perplexity scores from control and impaired subjects, along with their respective standard deviations.
Moreover, while in~\cite{cohen2020tale} the categorization is performed at the transcript level, our focus is on the categorization of subjects. 

%
\section{Background on Perplexity}\label{sec:background}
Most approaches rely on a simple yet powerful descriptive (and predictive) theoretical framework which is known as \textit{distributional hypothesis}.
The distributional hypothesis states that words that occur in similar contexts tend to convey similar meanings~\cite{harris1954distributional}.
%
%
Several techniques may devised to acquire the distributional profiles of terms, usually in the form of dense unit vectors of real numbers over a continuous, high-dimensional Euclidean space.
In this setting each word can be described through a vector, and each such vector can be mapped onto a multidimensional space where distance (such as, e.g., the Euclidean distance between vectors) acts like a proxy for similarity, and similarity can be interpreted as a metric. As a result, words with similar semantic content are expected to be closer than words semantically dissimilar. Different metrics can be envisaged, herein, to estimate the semantic proximity/distance of words and senses~\cite{colla2020novel}.
%
%

%
Language Models (LMs) are a statistical inference tool that allows estimating the probability of a word sequence $W = \{w_1, \dots, w_k\}$~\cite{manning1999foundations,goldberg2017neural}. 
%
Such probability can be computed as 
\begin{equation}\label{eq:w_seq_prob}
  p(W) = \prod_{i=1}^{k} p(w_{i}|w_{1},\dots,w_{i-1}), 
\end{equation}
which is customarily approximated as 
\begin{equation}\label{eq:w_seq_prob_approx}
  p(W) \approx \prod_{i=1}^{k} p(w_{i}|w_{i-N+1},w_{i-N+2},\dots,w_{i-1}).
\end{equation}
In the latter case only blocks of few (exactly $N$) words are considered to predict the whole $W$: we can thus predict the word sequence based on N-grams, that are blocks of two, three or four preceding elements (bi-grams, tri-grams, four-grams, respectively). 
In general N-gram models tend to obtain better performance as $N$ increases, with the drawback of making harder the estimation of $P(w_N|W_{1,N-1})$. Another issue featuring these models stems from the fact that when increasing the context size, it becomes less likely to find sequences with the same length in the training corpus. In order to deal with N-grams not occurring in the training corpus, called out-of-vocabulary N-grams, language models have to add an additional step of regularization to allow a non-zero probability to be associated to previously unseen N-grams~\cite{gale1994s,kneser1995improved}.
The probabilities assigned by language models are the result of a learning process, in which the model is exposed to a particular kind of textual data. The goal of the learning process is to train the model to predict word sequences that closely resemble the sentences seen during training.

As mentioned, LMs are basically probability distributions of word sequences: perplexity was originally conceived as an intrinsic evaluation tool for LMs, in that it can be used to measure how likely a given input sequence is, given a LM~\cite{goldberg2017neural}. 
%
This measure is defined as follows. Let us consider 
a word sequence of $k$ elements, $W = \{w_1, \dots, w_k\}$; since we are interested in evaluating the model on unseen 
data, the test sequence $W$ must be new, and not be part of the training set.
Given the language model $\text{LM}$, we can compute the probability of the sentence $W$, that is $\text{LM}(W)$. Such a probability would be a natural measure of the quality of the language model itself: the higher the probability, the better the model. The average log probability computed based on the model is defined as
\[
    \frac{1}{k} \log \prod_{i=1}^{k} \text{LM}(W) = \frac{1}{k} \sum_{i=1}^{k} \log\text{LM}(W),
\]
\noindent which amounts to the log probability of the whole test sequence $W$, divided by the number of tokens in sequence. 
The perplexity of sequence $W$ given the language model $\text{LM}$ is computed as 
%
%
%
%
\begin{equation}\label{eq:perplexity}
  \text{PPL}(\text{LM,W}) = \text{exp}\{-\frac{1}{k} \sum_{i=1}^{k} \log \text{LM}(w_i | w_{1:i-1})\}.
\end{equation}
%
%
%
It is now clear why low $\text{PPL}$ values (corresponding to high probability values) indicate that the word sequence fits well to the model or, equivalently, that the model is able to predict that sequence. 
Neural language models are language models based on neural networks. Such models improve on the language modeling capabilities of N-grams by exploiting the ability of neural networks to deal with longer histories. Additionally, neural models do not need regularization steps for unseen N-grams and address the data sparsity curse of N-grams by dealing with distributed representation. The predictive power of neural language models is higher than N-grams language models given the same training set. Despite the great improvement of neural language models on NLP tasks, these models are affected by training time higher than N-grams language models.

\section{Experiments}\label{sec:experiments}
The experimentation presented in this Section is concerned with answering one chief question: 
Whether the language of a specific class of subjects, diagnosed as suffering from disorders impacting on common linguistic abilities, can be automatically distinguished from that of healthy controls solely based on perplexity accounts. 
%
%
In this experiment we have used the Pitt Corpus, from which we selected the transcripts of responses to the Cookie Theft stimulus picture~\cite{goodglass1983boston}, 
which includes transcripts from patients with dementia diagnosis (n = $194$) and healthy controls (n = $99$).\footnote{The code for replicating the experiments is available at \url{https://github.com/davidecolla/semantic_coherence_markers},\cite{colla2022semantic_code}.}

\subsection{Compared LMs}
Different experimental setups have been designed in order to compare perplexity as computed by language models acquired by training with two different sorts of architectures: N-grams, and GPT-$2$. 

\subsubsection{N-grams}\label{subs:bigrams}
Since N-grams implement the simplest language model with context, where each word is conditioned on the preceding $N$-$1$ tokens only, we adopted N-grams for the first experimental setup. For the sake of clarity we introduce the formalization for Bigrams; such formulation can be further generalized to any $N$.

\noindent
We define the probability of a sequence of words $W_{1,n} = \{ w_1,w_2, \dots, w_n \}$ as:
\begin{equation*}
    P(W_{1,n}) = \prod_{i=1}^{n} P(w_i|w_{i-1}),
\end{equation*}
where the probability of each Bigram is estimated by exploiting the Maximum Likelihood Estimation (MLE)~\cite[Chap. 3]{jurafsky2014speech}.\footnote{In this setting, stopwords are customarily not filtered, as providing useful sequential information.} 
According to the MLE, we can estimate probability of the Bigram $(w_{i-1},w_i)$ as:
\begin{equation}\label{eq:bigrams_probability}
    P(w_i|w_{i-1}) = \frac{C(w_i|w_{i-1})}{C(w_{i-1})}
\end{equation}
where $C(w_i|w_{i-1})$ is the number of occurrences of the Bigram $(w_{i-1},w_i)$ in the training set, while $C(w_{i-1})$ counts the occurrences of the word $w_{i-1}$ only.
It is worth mentioning that training Bigrams on a limited vocabulary may lead to cases of out-of-vocabulary words, i.e., unseen words during the training process. Out-of-vocabulary words pose a problem in calculating the probability of the sentence in which they are involved: in such cases we are not able to compute the probability of the Bigram involving the unknown word, thus undermining the probability of the whole sequence.
%
%
We addressed the unseen N-grams issue through the interpolated Kneser-Ney Smoothing technique, which belongs to the family of interpolation strategies, and is based on the absolute discounting technique~\cite{kneser1995improved}. 
%
In the present setting we experimented with N-grams ranging  from $2$- to $5$-grams; the Kneser-Ney discounting factor $d$ was set to $0.1$.\footnote{To compute N-grams we exploited the Language Modeling Module (\textit{lm}) package from NLTK version $3.6.1$, \url{https://www.nltk.org/api/nltk.lm.html}. 
} 
The vocabulary was closed on each experiment: that is, the N-grams models employed in each experiment were acquired with the vocabulary obtained from the concatenation of the transcripts herein. %
Since the perplexity is bounded by the vocabulary size, fixing the cardinality of the vocabulary allows obtaining comparable perplexity scores from N-gram models trained across different corpora.


\subsubsection{GPT-$2$}
The second experimental setup that we designed exploits the GPT-$2$ neural model, in particular we used the GPT-$2$ pre-trained model available via the Hugging Face Transformers library.\footnote{\url{https://huggingface.co/gpt2}}
In this setting, the input text has been preprocessed by the pre-trained tokenizer and grouped into blocks of $1024$ tokens. The pre-trained model is specialized as Causal Language Model (CLM) on the input texts, that is, predicting a word given its left context.
Since the average log-likelihood for each token is returned as the loss of the model, the perplexity of a text is computed according to Equation~\ref{eq:perplexity}.

\subsection{Evaluation of the PPL-Based Categorizazion}\label{sub:experiment_3}
While the reliability associated to PPL has been extensively investigated in~\cite{colla2022semantic}, we presently investigate whether perplexity scores on the speech text transcripts allow discriminating patients from healthy controls. %
Publicly available data from the Pitt Corpus were used.\footnote{\url{https://dementia.talkbank.org/access/English/Pitt.html}.} These data were gathered as part of a larger protocol administered by the Alzheimer and Related Dementias Study at the University of Pittsburgh School of Medicine~\cite{becker1994natural}. In particular, we selected the descriptions provided to the Cookie Theft picture, which is a popular test used by speech-language pathologists to assess expository discourse in subjects with disorders such as dementia. 

\subsubsection{Materials}
The dataset is composed of $552$ files arranged into Control ($243$ items) and Dementia ($309$ items) directories. These correspond to multiple interviews to $99$ control subjects, and to $219$ subjects with dementia diagnosis.
%
%
Text documents herein were transcribed according to the CHAT format,\footnote{\url{https://talkbank.org/manuals/CHAT.pdf}.} so we pre-processed such documents to extract text. In so doing, the original text was to some extent simplified: e.g., pauses were disregarded, like hesitation phenomena, that were not consistently annotated~\cite{macwhinney2014childes,macwhinney2017tools}. 
To the ends of collecting enough text to be analyzed, we dropped the interviews of subjects that participated in only one interview.
We ended up with material relative to $74$ control subjects (for which overall $218$ transcripts were collected), and to $77$ subjects with dementia diagnosis (overall $192$ transcripts).

\begin{table}[!t]
\caption{Statistics describing the transcripts employed in Experiment $3$. For each class we report the average number of tokens per interview, the average number of unique tokens per interview, the number of participants, the overall number of transcripts and the type-token ratio (TTR).
}\label{tab:experiment_3_figures}
\begin{adjustbox}{width=\textwidth,center}
\begin{tabular}{l||c|c|c|c|c}
Class               & AVG Tokens & AVG Unique Tokens  & Participants & Transcripts & TTR \\ \hline
Control             & $437$         & $26$                  & $74$           & $218$          & $0.07$ \\
Alzheimer's Disease & $409$         & $25$                  & $77$           & $192$          & $0.08$ \\
\end{tabular}
\end{adjustbox}
\end{table}
The statistics describing number of tokens, number of unique tokens and type-token ratio for the transcripts employed in the Experiment 3 are presented in Table~\ref{tab:experiment_3_figures}.

\subsubsection{Procedure}
This experiment is aimed at testing the discriminative features of perplexity scores: more specifically, we tested a simple categorization algorithm to discriminate between mentally impaired and healthy subjects. 
We adopted the experimental setup from the work in~\cite{fritsch2019automatic}: two language models $LM_{C}$ and $LM_{AD}$ were acquired by employing all transcripts from Control and Alzheimer's disease groups, respectively. Such models are supposed to grasp the main linguistic traits of both groups speeches, thus representing the typical language adopted by subjects belonging to Control and AD classes.
For both groups we adopted a leave-one-subject-out setting, 
whereby 
language models were refined with files from all other subjects within the same group except for one, which was used for testing. For each subject $s$ we acquired the model $LM_s$ on the transcripts from the same group of $s$, except for those of the subject $s$.
Each transcript in the corpus was then characterized by two perplexity scores $P_C$ and $P_{AD}$, expressing the scores obtained through language models acquired on Control and AD groups, respectively. More precisely, if a subject $s$ was a member of the AD class, the scores $P_C$ for its transcripts were obtained through $LM_C$, while the scores $P_{AD}$ were computed by exploiting $LM_s$. Vice versa, if the subject $s$ was from the Control group, the scores $P_C$ for her/his transcripts were obtained through $LM_s$, while the scores $P_{AD}$ were computed by exploiting $LM_{AD}$.
Additionally, since we were interested in studying the scores featuring each subject, we synthesized the perplexity scores $P_C$ and $P_{AD}$ of each subject with the average of her/his transcripts scores, thus obtaining $\overline{P}_C$ and $\overline{P}_{AD}$. 

In order to discriminate AD patients from healthy subjects, we adopted a threshold-based classification strategy. 
Three different approaches were explored to estimate such threshold:
\begin{itemize}[noitemsep]
  \item [(i)] in the first setting we used the average perplexity scores characterizing all control subjects employed in the training process; 
  \item [(ii)] in the second setting we computed the threshold as the average perplexity score of all the subjects belonging to the AD class;
  \item [(iii)] in the third setting we estimated two different thresholds by exploiting the difference $\overline{P}_{AD} - \overline{P}_{C}$, by initially following the approach reported in~\cite{fritsch2019automatic} and~\cite{cohen2020tale}.
\end{itemize}
For each subject, the threshold estimation process was computed through a leave-one-subject-out setting, and repeated for the three approaches from (i) to (iii).
In the first setting the threshold was estimated on all the subjects from the control group except for the test subject $s$: 
for each subject $s$ we computed the threshold as the average of $\overline{P}_C$ scores for all subjects in the control group except for $s$ ---if $s$ was from the healthy controls group---. In case the perplexity score $\overline{P}_C$ for the subject $s$ was higher than the 
healthy controls threshold, we marked the subject as suffering from AD; as healthy otherwise.
Similarly, in the second setting we computed the threshold as the average of $\overline{P}_{AD}$ scores for all subjects in the AD group except for $s$. In case the perplexity score $\overline{P}_{AD}$ for the subject $s$ was higher than the average of AD class threshold, we marked the subject as healthy; as suffering from AD otherwise.
The rationale underlying the first two settings is that each subject may be characterized more accurately by LMs acquired on transcript from the same group: in other words, we expected lower perplexity scores to be associated to 
control (AD) subjects, rather than subjects belonging to the other class, with LMs trained or fine-tuned on transcripts from control (AD) subjects. %

Following the literature, in the third setting we characterized each subject with the difference $D = \overline{P}_{AD} - \overline{P}_{C}$. 
We defined two 
thresholds, $\overline{D}_{AD}$ which was computed as the average of all the difference scores from patients in the AD group and $\overline{D}_{C}$, defined as the average of all the difference scores from healthy controls. In both cases we considered all the patients belonging to the group except for the test subject $s$ ($s$ was held out with the only purpose to rule out her/his contribution from $\overline{D}_{AD}$ or $\overline{D}_{C}$). 
Different from literature ---where equal error rate is used---, we employ $\overline{D}_{AD}$ and $\overline{D}_{C}$ as compact descriptors for the classes $AD$ and $C$, respectively. The rationale underlying this categorization schema is that a subject is associated to the class that exhibits most similar perplexity score to her/his own.
We categorize a subject $s$ by choosing the class associated to the threshold (either $\overline{D}_{AD}$ or $\overline{D}_{C}$) featured by smallest margin with the $D$ value associated to the subject $s$, according to the following formula: 
\begin{equation}\label{eq:d_average}
\mathrm{class}(s) = \underset{x \in \{C, AD\}}{\mathrm{argmin}}
    \left| \, D - \overline{D}_{x}  \, \right|.   
\end{equation}
This setting (involving $\overline{D}_{AD}$ and $\overline{D}_{C}$) will be referred to as $\overline{D}$.

Furthermore, we refined the decision rule $\overline{D}$ to account for standard deviation information.
Together with the average $\overline{D}_{AD}$ and $\overline{D}_{C}$, we computed also $\sigma_{AD}$ and $\sigma_{C}$ as the standard deviations of the difference scores $D$ for impaired and control groups. 
We explored the $3\sigma$ rule, which is a popular heuristic in empirical sciences: it states that in populations that are assumed to be described by a normally distributed random variable, over $99.7\%$ values lie within three standard deviations of the mean, $95.5\%$ within two standard deviations, and $68.3\%$ within one standard deviation~\cite{helms2009mathematics}.
On this basis we explored the three options by adding $1$, $2$ and $3$ standard deviations to average scores: the best results were obtained by employing $2$ standard deviations. Our thresholds were then refined as follows: 
\begin{eqnarray*}\label{eq:Dstar}
  \overline{D}^*_{AD} & = & \overline{D}_{AD} + 2 \cdot \sigma_{AD}\text{, and}\\
  \overline{D}^*_{C} & = & \overline{D}_{C} - 2 \cdot \sigma_{C}.
\end{eqnarray*}
The updated decision rule for categorization was then reshaped as
\begin{equation}\label{eq:d_star_decision_rule}
\mathrm{class}(s) = \underset{x \in \{C, AD\}}{\mathrm{argmin}}
    \left| \, D -\overline{D}^{*}_{x}  \, \right|.
\end{equation}
This setting, involving $\overline{D}^*_{AD}$ and $\overline{D}^*_{C}$, will be referred to as $\overline{D}^*$.

A twofold experimental setting has been devised, including experiments with N-grams and GPT-$2$, adopting a window size set to $20$ in order to handle shorter text samples (the shortest text in the training data contains only $23$ tokens). In the case of N-grams, the models were acquired for $2$-grams to $5$-grams; the GPT-$2$ model was fine-tuned employing $5$, $10$, $20$ and $30$ epochs.

\subsubsection{Evaluation Metrics}\label{subs:evaluation_metrics}
To evaluate the results we adopted the Precision and Recall metrics (specificity and sensitivity) along with their harmonic mean, F1 score, and accuracy.
Precision (specificity) is defined as 
$P = \frac{TP}{TP+FP}$,
while Recall (sensitivity) is defined as 
$R = \frac{TP}{TP+FN}$. 
While precision provides an estimation of how precise a categorization system is, recall indicates how many results were identified out of all the possible ones.
$F_1$ measure is then used to provide a synthetic value of Precision and Recall, whereby the two measures are evenly weighted through their harmonic mean:
    $F_1 = 2 \cdot \frac{P \cdot R}{P+R}$

Accuracy was computed as $ACC = \frac{TP+TN}{P+N}$, that is as the fraction of correct predictions (the sum of TP and TN) over the total number of records examined (the sum of positives and negatives, P and N).

Finally, in order to record a synthetic index to assess accuracy and F1 scores on the two groups at stake, we used the harmonic mean among these three values. It was computed as 
\[
\text{HM}(\text{Acc.}, \text{F1}_{AD}, \text{F1}_{C}) = 
{\frac {n}{\sum \limits _{i=1}^{n}{\frac {1}{x_{i}}}}} =
\left({\frac {\sum \limits _{i=1}^{n}x_{i}^{-1}}{n}}\right)^{-1}.
\]
where $n$ was set to the number of $x_{i}$ values being averaged.

\subsubsection{Results}

\begin{figure}[t]
     \centering
     \includegraphics[width=0.65\columnwidth]{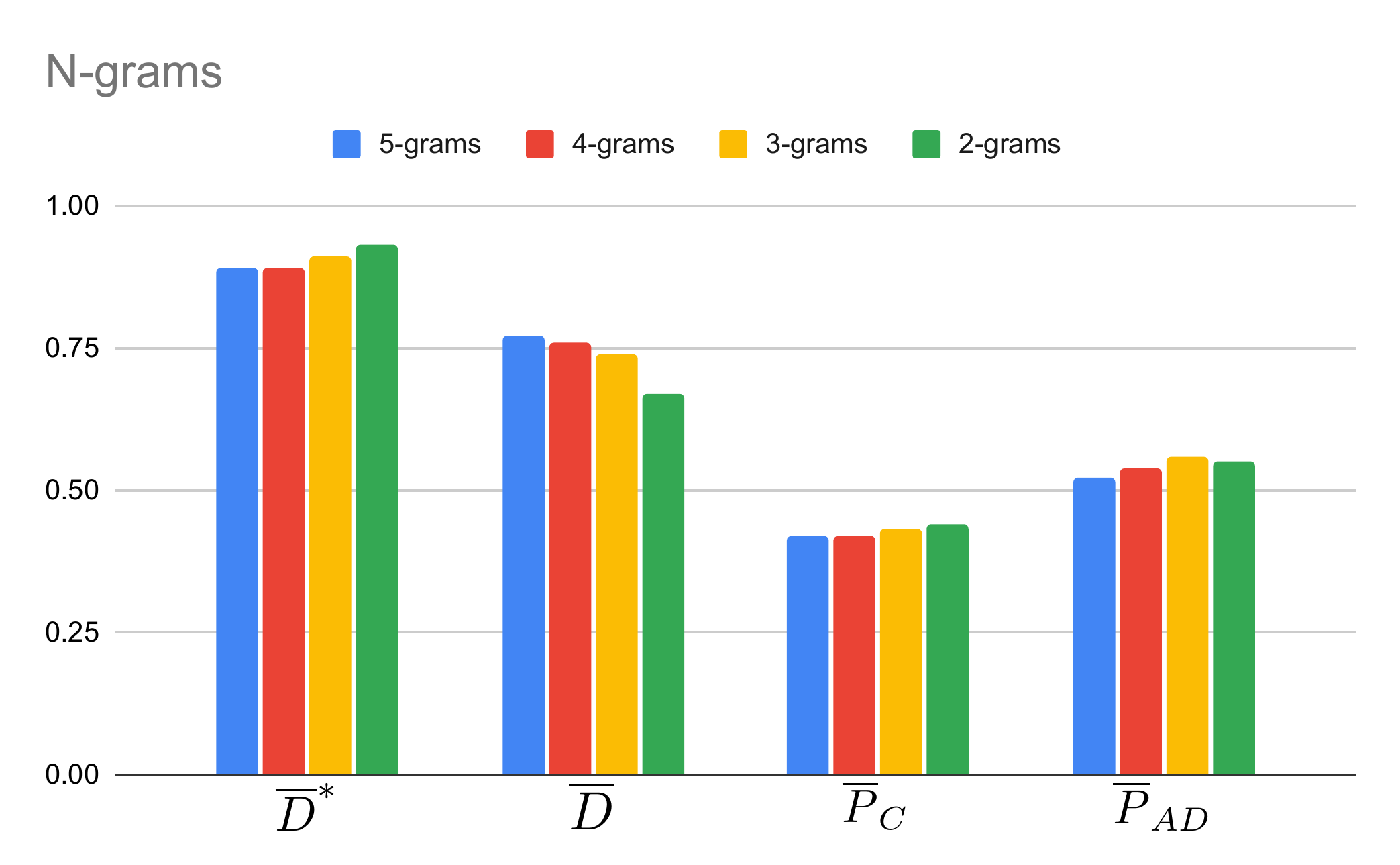}\hfill
     \includegraphics[width=0.65\columnwidth]{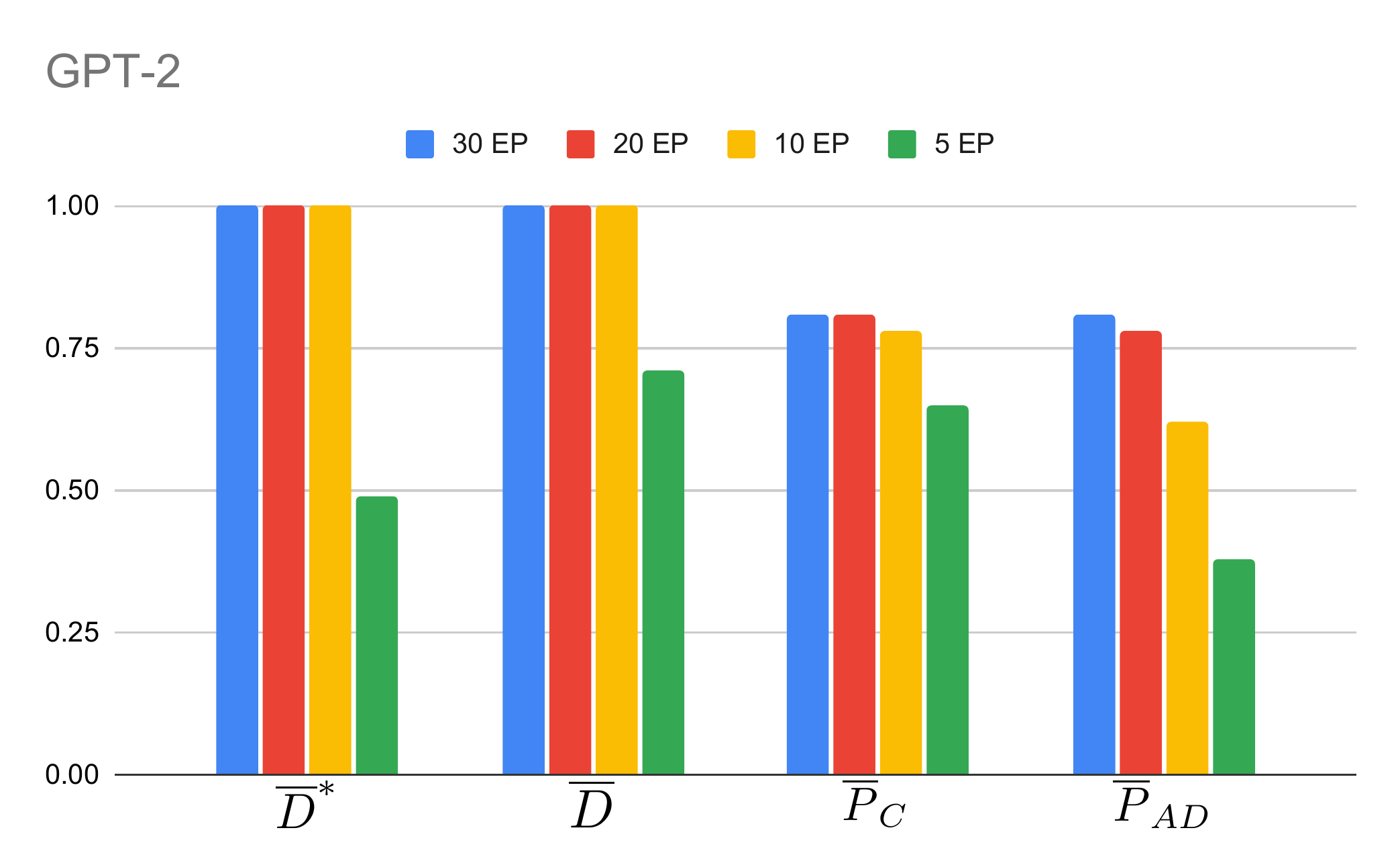}
     \caption{Plot of the accuracy scores for the third experiment on the categorization of AD/control subjects. The histograms in the top sub-figure show the accuracy on N-grams, while the histograms at the bottom report results obtained through GPT-2 models. Different colors correspond to N-gram of differing order and to different fine-tuning epochs, respectively. The histograms illustrate the scores obtained through $\overline{D}^*$, $\overline{D}$, $\overline{P}_{C}$ and $\overline{P}_{AD}$ decision rules, respectively.}
     \label{fig:exp3_acc}
 \end{figure}

The overall accuracy scores are presented in Figure~\ref{fig:exp3_acc}, while detailed figures across different experimental conditions are presented in  Table~\ref{tab:experiment_3_detailed_results}, in~\ref{appendix:materials}.

Let us start by reporting the results from N-gram models. The overall most effective strategy is 
$\overline{D}^{*}$ (Eq.~\ref{eq:d_star_decision_rule}), based on a threshold using the difference between AD patients and healthy controls, extended with the $3\sigma$ rule.
The best performing model is based on Bigrams, and obtained $.93$ accuracy, $.92$ F1 score on the AD class, and $.93$ F1 score on the C class. 
The models employing PPL scores from the control group (indicated as $P_{C}$ in Figure~\ref{fig:exp3_acc} and in  Table~\ref{tab:experiment_3_detailed_results}) obtained the lowest accuracy scores in all conditions, well below the random guess, while the accuracy yielded by the $\overline{P}_{AD}$ strategy is always above $.5$.
In general we observe that increasing the length of the Markovian assumption reduces the accuracy of N-gram models for all decision rules (employing more context seems to be slightly detrimental for such models), with the exception of the $\overline{D}$ strategy. 

The results obtained by the GPT-2 models reveal overall higher accuracy, ranging from $.71$ for the best model acquired with $5$ epochs of fine-tuning to $1.00$ for all further fine-tuning steps. The same profile describes the F1 scores recorded on the sub-tasks focused on AD and control subjects, respectively, varying from around $0.69$ for the best model acquired with $5$ epochs of fine-tuning ($\overline{D}$ strategy on the AD class) to $1.00$ for all other models and sub-tasks. 
If we consider the efficacy of thresholding strategies and associated decision rules, the refined difference rule $\overline{D}$ is the best performing strategy for GTP-2 based models, as witnessed by the rightmost column in Table~\ref{tab:experiment_3_detailed_results}.
Such scores report the harmonic mean among accuracy, F1 score on categorization of AD subjects and on categorization of control subjects. A compact view on data from the same column is provided in Table~\ref{tab:mean_HM}, illustrating the best strategy for each model at stake.

\begin{table}[]
\centering
\caption{Study to compare the effectiveness of the thresholding and categorization strategies for each LM. The top scoring strategy is reported for each model.}
\label{tab:mean_HM}
\begin{adjustbox}{center}
%
\begin{tabular}{c|c|c}
N-gram models   &  categorization strategy        &   mean HM score \\\hline
    2-grams         & $\overline{D}^{*}$    & $0.93$ \\
    3-grams         & $\overline{D}^{*}$    & $0.91$ \\
    4-grams         & $\overline{D}^{*}$    & $0.89$ \\
    5-grams         & $\overline{D}^{*}$    & $0.89$ \\ \hline
    GPT-2 models: epochs  &  categorization strategy            &   mean HM score \\\hline
    5  epochs       & $\overline{D}$                  &   $0.71 $ \\
    10 epochs       & $\overline{D},\overline{D}^{*}$ &  $1.00 $ \\
    20 epochs       & $\overline{D},\overline{D}^{*}$ & $1.00 $ \\
    30 epochs       & $\overline{D},\overline{D}^{*}$ & $1.00 $ \\
\end{tabular}
\end{adjustbox}
\end{table}

To frame our results with respect to literature, let us start from the accuracy of the baseline clinical diagnosis obtained in the first version of the study by Becker and Colleagues~\cite{becker1994natural}: it was $86\%$, and after considering follow-up clinical data this datum raised to $91.4\%$, with a $0.988$ sensivity and $0.983$ specificity. This is what subsequent literature considered as the gold standard against which to compare experimental outputs. 
We recall that such data are particularly relevant as human evaluation included various analytical steps, such as medical and neurologic history and examination, semistructured psychiatric interview, and neuropsychological assessments. 
Experimental results provided in subsequent work 
approach those ratings by employing solely transcripts of descriptions to a rather simple picture.
A relevant work attained $85.6\%$ accuracy through LSTM based models~\cite{fritsch2019automatic} in the categorization of individual transcripts. Such results were then replicated and improved in the work by~\cite{cohen2020tale}, where the best reported model experimentally obtained a $0.872$ accuracy.

\subsection*{General Discussion}
Provided that our experimental results seem to outperform the accuracy scores reported in literature, 
%
we realized that a short, controlled elicitation task can potentially outperform natural linguistic data obtained from speakers. 
The quality of our results needs be checked in different settings (further languages, varied experimental conditions: much experimental work thus still needs to be done), but this fact provides evidence that specialists may be effectively assisted by systems employing a technology based on language models and perplexity scores.
Also, by comparing language models as different as N-grams and models based on the more recent GPT-2, we observed that Bigrams outperform a GPT-2 model fine tuned for $5$ epochs. This fact may provide insights on the possible trade-off between accuracy of the results and computation time and costs.

While perplexity proved to be overall a viable tool to investigate human language, we found consistent differences in the outputs of the models at stake, mostly stemming from intrinsic properties of the LMs, from the amount of context considered by the models, from the size of available training data, and from the amount of training employed to refine models themselves. 
One first datum is that even though N-grams can be hardly compared to
GPT-2-based models, nonetheless it may be helpful trying to discern the scenarios in which such models provide better results.
%
%
%
It was somehow surprising that in our experiment the accuracy level attained by the best-performing N-gram model (2-grams) achieved a $0.93$ harmonic mean improving on the best GPT-2-based model (HM=$0.73$; please refer to Table~\ref{tab:experiment_3_detailed_results}), fine tuned for $5$ epochs and employing the $\overline{D}$ decision rule. 

This result may be understood in the light of the rather regular language used for the descriptions to the Cookie Theft picture, that thereby turned out to be less demanding for the N-gram LMs. In these respects, a lesson learned is that N-grams can be employed in scenarios where the task is less difficult on lexical and linguistic accounts: 
in some instances of such problems adopting N-gram models may be convenient (considering both training and testing efforts) with respect to the more complete and computationally expensive Transformer models.
Few data may be useful to complete this note on the trade-off between accuracy and computational effort.
Our experiments were performed on machinery provided by the Competence Centre for Scientific Computing~\cite{aldinucci2017occam}. In particular, we exploited nodes with 2x Intel Xeon Processor E5-2680 v3 and 128GB memory. 
Reported experiments took around $8$ hours for each GPT-$2$ setting and about $12$ minutes for all the N-gram models.

\section{Conclusions}
The study reported in this work explored how suited perplexity is 
to support  automatic linguistic analysis for clinical diagnoses.
The diagnosis of dementia is a complex process that is long and labor intensive, involving a neuropsychiatric evaluation that includes medical and neurologic history and examination, semistructured psychiatric interview, and neuropsychological assessments~\cite{huff1987cognitive,lopez1990reliability}. 
Being able to define a linguistic marker to detect symptoms of mental disorders would thus provide clinicians with automatic procedures for language analysis that can contribute to the early diagnosis and treatment of mental illnesses in an efficient and noninvasive fashion. %
We thus addressed one basic research issue: 
whether and to what extent perplexity scores allow categorizing transcripts of healthy subjects and subjects suffering from Alzheimer Disease (AD). In this experiment we used a publicly available dataset, the Pitt Corpus.
A widely varied experimental setting was designed to investigate the predictive and discriminative power of perplexity scores, and to assess how the resulting categorization accuracy varies in function of the amount of training/fine-tuning employed to acquire the LMs. We compared (2, 3, 4 and 5) N-gram models, $0$ to $30$ (GPT-2) fine-tuning epochs, and four different thresholding strategies, as well. Novel thresholds
were proposed, and compared to those
reported in literature: the newly proposed categorization strategies ensure consistent improvement over state-of-the-art results. 
%
%
%

A final remark relates to an outlook on future work. 
Different language models can attain  results possibly featured by analogous accuracy with a fraction of training/fine-tuning efforts:
e.g., we conducted preliminary tests, not reported here for brevity, also on LSTMs that revealed poor performance, paired with a computational load higher than for the GPT-$2$ architecture.\footnote{More specifically, perplexity scores computed through LSTMs were highly volatile (with standard deviation values often overcoming mean perplexity values), 
even increasing the number of training epochs, which required almost twice the time necessary to train the GPT-2 base model.}
%
%
%
%
%
Also, different categorization algorithms may be adopted to discriminate patients from control subjects; 
refinements to both employed LMs and overall categorization strategy may result in substantial improvements. Yet, further experiments are needed to assess perplexity on larger samples, and on different sorts of spoken language: as mentioned, the language required to comment the Cookie Theft picture is quite a regular one. A richer, fuller characterization of the discriminative power of perplexity scores will involve experimenting also on different languages, and the associated language models.
 
However, the findings from this proof-of-concept study have several implications: 
while predicting whether the author of a transcript was afflicted by dementia or a healthy subject, we obtained valuable results, especially if we consider that our predictions were based solely on perplexity scores, with a substantial reduction in the amount of information with respect to the clinical evidence collected all throughout the diagnosis steps employed by human experts to face the same categorization task~\cite{becker1994natural}. 
%
%
%


\bibliography{0_main}

\appendix
\newpage

\section{Detailed results}\label{appendix:materials}

\begin{table}[h]
\centering
\caption{Detailed results. The table reports Accuracy (Acc.), Precision (P), Recall (R) and F1 scores on both tasks aimed at identifying AD and Control subjects. The rightmost column reports the harmonic mean (HM) of the accuracy, F1 score on the AD and C classes. Best results are marked in boldface.}\label{tab:experiment_3_detailed_results}
\vspace{4mm}
\begin{adjustbox}{width=.89\textwidth,center}
\begin{tabular}{ll||c|ccc||ccc||c}
 \multicolumn{2}{c||}{\multirow{2}{*}{Model}} & \multirow{2}{*}{Acc.} & \multicolumn{3}{c||}{Dementia (AD)} & \multicolumn{3}{c||}{Control (C)}& \multirow{2}{*}{HM(acc,F1$_{AD}$,F1$_{C}$)}\\ \cline{4-9}
 & & & P & R & F1 & P & R & F1 & \\ \hline
 \multirow{3}{*}{2-grams} 
  & $\overline{P}_{C}$     & $0.44$ & $0.41$ & $0.21$ & $0.28$ & $0.46$ & $0.69$ & $0.55$ & $0.39$ \\
  & $\overline{P}_{AD}$    & $0.55$ & $0.54$ & $0.75$ & $0.63$ & $0.57$ & $0.34$ & $0.42$ & $0.52$ \\
  & $\overline{D}$         & $0.67$ & $0.68$ & $0.66$ & $0.67$ & $0.66$ & $0.68$ & $0.67$ & $0.67$ \\
  & $\overline{D}^*$       & $\textbf{0.93}$ & $0.99$ & $0.87$ & $\textbf{0.92}$ & $0.88$ & $0.99$ & $\textbf{0.93}$ & $\textbf{0.93}$ \\ \hline
  \multirow{3}{*}{3-grams}  
  & $\overline{P}_{C}$     & $0.43$ & $0.40$ & $0.22$ & $0.28$ & $0.44$ & $0.65$ & $0.53$ & $0.39$ \\
  & $\overline{P}_{AD}$    & $0.56$ & $0.55$ & $0.70$ & $0.62$ & $0.57$ & $0.41$ & $0.47$ & $0.54$ \\
  & $\overline{D}$         & $0.74$ & $0.76$ & $0.71$ & $0.74$ & $0.72$ & $0.77$ & $0.75$ & $0.74$ \\
  & $\overline{D}^*$       & $\textbf{0.91}$ & $1.00$ & $0.83$ & $\textbf{0.91}$ & $0.85$ & $1.00$ & $\textbf{0.92}$ & $\textbf{0.91}$ \\ \hline
  \multirow{3}{*}{4-grams}  
  & $\overline{P}_{C}$     & $0.42$ & $0.38$ & $0.23$ & $0.29$ & $0.43$ & $0.61$ & $0.51$ & $0.38$ \\
  & $\overline{P}_{AD}$    & $0.54$ & $0.54$ & $0.65$ & $0.59$ & $0.54$ & $0.43$ & $0.48$ & $0.53$ \\
  & $\overline{D}$         & $0.76$ & $0.81$ & $0.70$ & $0.75$ & $0.73$ & $0.82$ & $0.77$ & $0.76$ \\
  & $\overline{D}^*$       & $\textbf{0.89}$ & $1.00$ & $0.78$ & $\textbf{0.88}$ & $0.81$ & $1.00$ & $\textbf{0.90}$ & $\textbf{0.89}$ \\ \hline
  \multirow{3}{*}{5-grams}  
 
  & $\overline{P}_{C}$     & $0.42$ & $0.38$ & $0.23$ & $0.29$ & $0.43$ & $0.61$ & $0.51$ & $0.38$ \\
  & $\overline{P}_{AD}$    & $0.52$ & $0.53$ & $0.62$ & $0.57$ & $0.52$ & $0.42$ & $0.46$ & $0.52$ \\
  & $\overline{D}$         & $0.77$ & $0.86$ & $0.66$ & $0.75$ & $0.72$ & $0.89$ & $0.80$ & $0.77$ \\
  & $\overline{D}^*$       & $\textbf{0.89}$ & $1.00$ & $0.79$ & $\textbf{0.88}$ & $0.82$ & $1.00$ & $\textbf{0.90}$ & $\textbf{0.89}$ \\ \midrule
  \multirow{3}{*}{GPT-$2$ $5$ epochs}   
  & $\overline{P}_{C}$     & $0.65$ & $0.64$ & $0.70$ & $0.67$ & $0.66$ & $0.59$ & $0.62$ & $0.65$ \\
  & $\overline{P}_{AD}$    & $0.38$ & $0.42$ & $0.58$ & $0.49$ & $0.29$ & $0.18$ & $0.22$ & $0.33$ \\
  & $\overline{D}$         & $\textbf{0.71}$ & $0.76$ & $0.62$ & $\textbf{0.69}$ & $0.67$ & $0.80$ & $\textbf{0.73}$ & $\textbf{0.71}$ \\
  & $\overline{D}^*$       & $0.49$ & $0.50$ & $0.09$ & $0.15$ & $0.49$ & $0.91$ & $0.64$ & $0.30$ \\ \hline
  \multirow{3}{*}{GPT-$2$ $10$ epochs}  
  & $\overline{P}_{C}$     & $0.78$ & $0.70$ & $0.99$ & $0.82$ & $0.98$ & $0.57$ & $0.72$ & $0.77$ \\
  & $\overline{P}_{AD}$    & $0.62$ & $0.63$ & $0.58$ & $0.61$ & $0.60$ & $0.65$ & $0.62$ & $0.62$ \\
  & $\overline{D}$         & $\textbf{1.00}$ & $1.00$ & $1.00$ & $\textbf{1.00}$ & $1.00$ & $1.00$ & $\textbf{1.00}$ & $\textbf{1.00}$ \\
  & $\overline{D}^*$       & $\textbf{1.00}$ & $1.00$ & $1.00$ & $\textbf{1.00}$ & $1.00$ & $1.00$ & $\textbf{1.00}$ & $\textbf{1.00}$ \\ \hline
  \multirow{3}{*}{GPT-$2$ $20$ epochs}  
  & $\overline{P}_{C}$     & $0.81$ & $0.73$ & $1.00$ & $0.84$ & $1.00$ & $0.61$ & $0.76$ & $0.80$ \\
  & $\overline{P}_{AD}$    & $0.78$ & $0.91$ & $0.64$ & $0.75$ & $0.71$ & $0.93$ & $0.81$ & $0.78$ \\
  & $\overline{D}$         & $\textbf{1.00}$ & $1.00$ & $1.00$ & $\textbf{1.00}$ & $1.00$ & $1.00$ & $\textbf{1.00}$ & $\textbf{1.00}$ \\
  & $\overline{D}^*$       & $\textbf{1.00}$ & $1.00$ & $1.00$ & $\textbf{1.00}$ & $1.00$ & $1.00$ & $\textbf{1.00}$ & $\textbf{1.00}$ \\ \hline
  \multirow{3}{*}{GPT-$2$ $30$ epochs}  
  & $\overline{P}_{C}$     & $0.81$ & $0.73$ & $1.00$ & $0.84$ & $1.00$ & $0.61$ & $0.76$ & $0.80$ \\
  & $\overline{P}_{AD}$    & $0.81$ & $0.96$ & $0.65$ & $0.78$ & $0.73$ & $0.97$ & $0.83$ & $0.80$ \\
  & $\overline{D}$         & $\textbf{1.00}$ & $1.00$ & $1.00$ & $\textbf{1.00}$ & $1.00$ & $1.00$ & $\textbf{1.00}$ & $\textbf{1.00}$ \\
  & $\overline{D}^*$       & $\textbf{1.00}$ & $1.00$ & $1.00$ & $\textbf{1.00}$ & $1.00$ & $1.00$ & $\textbf{1.00}$ & $\textbf{1.00}$ \\

\end{tabular}
\end{adjustbox}
\end{table}

\end{document}